\documentclass[lettersize,journal]{IEEEtran}
\usepackage{amsmath,amsfonts}
\usepackage{algorithmic}
\usepackage{array}
\usepackage[caption=false,font=normalsize,labelfont=sf,textfont=sf]{subfig}
\usepackage{textcomp}
\usepackage{stfloats}
\usepackage{url}
\usepackage{verbatim}
\usepackage{graphicx}
\usepackage{xcolor} 
\usepackage{ulem} 
\usepackage{orcidlink}

\def\BibTeX{{\rm B\kern-.05em{\sc i\kern-.025em b}\kern-.08em
    T\kern-.1667em\lower.7ex\hbox{E}\kern-.125emX}}
\usepackage{balance}

\usepackage{tabularx}
\usepackage{threeparttable}
\usepackage{multirow}
\newcolumntype{Y}{>{\centering\arraybackslash}X}

\newcommand{\deleted}[1]{}
\newcommand{\added}[1]{{#1}}

\newcommand{\reducespace}{\vspace{-4mm}}
\newcommand{\reducespacebeforesection}{\vspace{-0mm}}

\begin{document}
\title{PRIOT: Pruning-Based Integer-Only\\Transfer Learning for Embedded Systems}

\markboth{}{PRIOT: Pruning-Based Integer-Only Transfer Learning for Embedded Systems}

\author{Honoka Anada\raisebox{0.5ex}{\orcidlink{0009-0005-2695-6734}}, Sefutsu Ryu, Masayuki Usui, Tatsuya Kaneko\raisebox{0.5ex}{\orcidlink{0000-0002-0951-875X}},\\and Shinya Takamaeda-Yamazaki\raisebox{0.5ex}{\orcidlink{0000-0003-3441-1695}}, \textit{Member, IEEE}
\thanks{The authors are with the Graduate School of Information Science and Technology, The University of Tokyo, Tokyo 113-8656, Japan (e-mail: honokaanada@is.s.u-tokyo.ac.jp; ryusef@is.s.u-tokyo.ac.jp; mu2519@is.s.u-tokyo.ac.jp; tatsuya-kaneko@is.s.u-tokyo.ac.jp; shinya@is.s.u-tokyo.ac.jp). \vspace{2mm}\\\copyright 2025 IEEE.  Personal use of this material is permitted.  Permission from IEEE must be obtained for all other uses, in any current or future media, including reprinting/republishing this material for advertising or promotional purposes, creating new collective works, for resale or redistribution to servers or lists, or reuse of any copyrighted component of this work in other works.}}

\maketitle

\begin{abstract}
On-device transfer learning is crucial for adapting a common backbone model to the unique environment of each edge device. Tiny microcontrollers, such as the Raspberry Pi Pico, are key targets for on-device learning but often lack floating-point units, necessitating integer-only training. Dynamic computation of quantization scale factors, which is adopted in former studies, incurs high computational costs. Therefore, this study focuses on integer-only training with static scale factors, which is challenging with existing training methods. We propose a new training method named PRIOT, which optimizes the network by pruning selected edges rather than updating weights, allowing effective training with static scale factors. The pruning pattern is determined by the edge-popup algorithm, which trains a parameter named score assigned to each edge instead of the original parameters and prunes the edges with low scores before inference. Additionally, we introduce a memory-efficient variant, PRIOT-S, which only assigns scores to a \deleted{randomly selected few }\added{small fraction of }edges. We implement PRIOT and PRIOT-S on the Raspberry Pi Pico and evaluate their accuracy and computational costs \deleted{using a tiny CNN model and the rotated MNIST dataset }\added{using a tiny CNN model on the rotated MNIST dataset and the VGG11 model on the rotated CIFAR-10 dataset}. Our results demonstrate that PRIOT improves accuracy by 8.08 to 33.75 percentage points over existing methods, while PRIOT-S reduces memory footprint with minimal accuracy loss.
\end{abstract}


\reducespacebeforesection
\section{Introduction}\label{sec/introduction}
\IEEEPARstart{O}{n-device} training and inference of neural networks are currently becoming increasingly important owing to the rising significance of deep learning across various fields and the expanding volumes of communications. In particular, on-device transfer learning on low-end edge devices is crucial for adapting a model trained on a central server to the specific environment of each device after distribution\added{, with various potential applications including anomaly detection on IoT devices and health monitoring on wearable devices.}

Tiny microcontrollers, such as the Raspberry Pi Pico, are important targets for edge computing due to their affordability and extreme power efficiency. Hence, on-device transfer learning on those devices is also crucial. However, they sometimes do not have floating-point units (FPUs). Floating-point arithmetic on a device without FPUs requires software emulation, which incurs extremely high computational costs. Therefore, in this study, we aim to represent all weights, activations, and gradients in integers, specifically 8-bit integers, and perform the entire training using only integer arithmetic.

Several studies have investigated integer-only training for neural networks, but they dynamically compute the quantization scale factors during training \cite{iclr18_wage}\cite{tpds22_niti}. Here, \textit{scale factor} refers to the amount of right-shifting a large-bit-width multiply-accumulate result into a small bit-width integer. However, this dynamic scaling poses significant challenges for lightweight computing on tiny devices, including increased memory footprint during both training and inference. Therefore, this study focuses on static-scale integer-only training, where all scale factors are fixed during training and inference. We discovered that existing integer-only training methods struggle with this approach, experiencing training collapse in the middle and
resulting in significantly low accuracy.

\begin{figure}
    \centering
    \includegraphics[width=\linewidth]{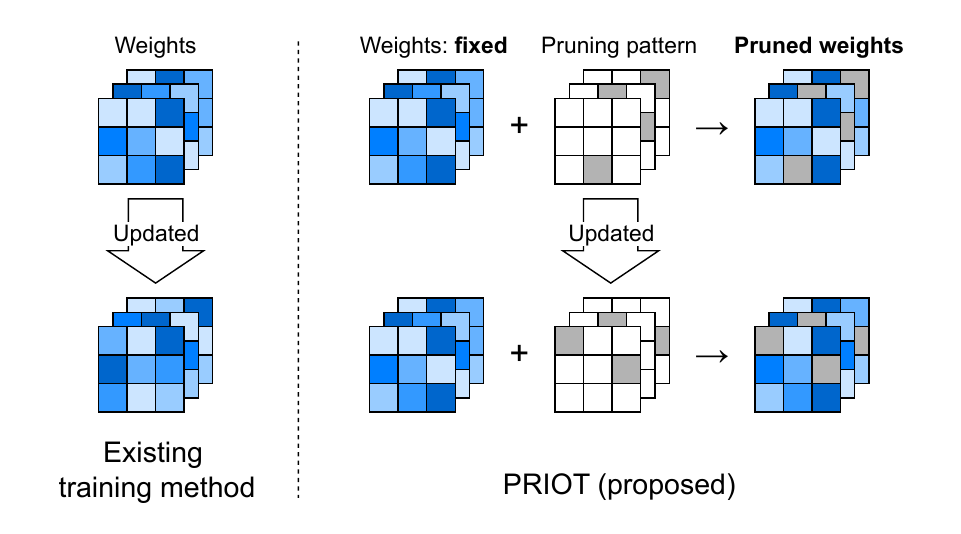}
    \caption{Overview of PRIOT. PRIOT trains the model by pruning the edges rather than updating the weights.}
    \label{fig/priot}
    \reducespace
\end{figure}

To overcome this challenge, we propose an alternative training method named PRIOT in this study. Figure \ref{fig/priot} illustrates the overview of PRIOT. PRIOT trains the model by \textit{pruning} the pre-trained edges rather than updating their weights. In PRIOT, the pruning-only approach ensures that the activation distributions remain stable throughout the training, preventing training collapse. For the pruning pattern training in PRIOT, we employ the edge-popup algorithm \cite{cvpr20_strong_lth}. This algorithm assigns a score to each edge, updates the scores instead of weights by backpropagation during training, and prunes edges with low scores before inference. In this study, this algorithm is performed with integer arithmetic only, along with a few modifications such as using pre-trained weights instead of randomly initialized weights.

Unfortunately, despite its effectiveness in integer-only training, PRIOT requires a greater memory footprint than ordinary training because of its additional storage requirements for the scores. Therefore, we also propose PRIOT-S, a memory-saving variant of PRIOT, designed for compatibility with memory-limited situations. PRIOT-S assigns scores to \deleted{a randomly selected few }\added{a small fraction of }edges to save the memory occupied by the scores. 

We implement PRIOT and PRIOT-S for the Raspberry Pi Pico, a tiny microcontroller without FPUs. We evaluate the accuracies, training times, and memory footprints of our proposed methods using a tiny CNN model and the rotated MNIST dataset, \added{alongside the additional accuracy evaluation on the rotated CIFAR-10 dataset using the VGG11 model.} Consequently, PRIOT achieves an accuracy improvement in the range of 8.08 to 33.75 percentage points over the existing integer-only training method using static scale factors, demonstrating that it overcomes the challenges of static-scale integer-only training. While PRIOT-S exhibits smaller accuracy improvement compared to PRIOT, it significantly reduces the memory footprint compared to PRIOT. Hence, PRIOT-S is effective at reducing computational costs when a certain level of accuracy loss is acceptable.

\reducespacebeforesection
\section{Background}\label{sec/background}
\subsection{Related Work}
On-device learning on microcontrollers has been investigated by many studies, and some of them are targeted at extremely small devices, such as Raspberry Pi Pico\cite{pico-matsutani}\cite{pruningforpower} and Arduino Nano\cite{odl-look-at-fl}. However, most of them employ floating-point arithmetic, suffering from high computational costs and limitations in model size. A few studies have investigated integer-only training on microcontrollers, such as \cite{octo} and \cite{256kbmemory}. Specifically, \cite{256kbmemory} enabled neural network training with 256KB of static RAM (SRAM), supported by their proposed sparse update method. However, this study uses dynamic scale factor computation, which brings several challenges in lightweight computation, as described in the next section.

\vspace{-1mm}
\subsection{Challenges in Integer-Only Training}
Integer-only training is a promising approach for reducing computational costs of neural network training, which some former studies have explored. WAGE\cite{iclr18_wage} was the first study to use 8-bit integer values for all weights, activations, gradients, and errors during training. While WAGE used floating-point numbers for the first and last layers and cross-entropy backward computation, a later study named NITI\cite{tpds22_niti} replaced them with integer arithmetic. 

In integer-only training methods, all weights, activations, and gradients are represented by integers, typically 8-bit integers. Since the output of the matrix multiplication between two 8-bit integer tensors results in a 32-bit-integer tensor $x_\text{int32}$, we need to right-shift the elements by a suitable scale factor $s$ to convert them to an 8-bit-integer tensor $x_\text{int8}$. Existing integer-only training methods, including WAGE and NITI, dynamically determine $s$ by examining the computed $x_\text{int32}$ values.

However, this dynamic scaling is inappropriate for lightweight computing on tiny devices for several reasons. \underline{First}, this approach requires storing all values of $x_\text{int32}$ once since $s$ is determined after examining all values of $x_\text{int32}$. The increase in memory footprint caused by this is critical for tiny devices. \underline{Second}, the dynamic computation of scale factors is quite complicated in the model with bias parameters or skip connections, both of which are common in recent neural network architectures. \underline{Finally}, dynamic scaling is also required during inference with this approach, increasing the computational costs of on-device inference as well. 

\begin{figure}
    \centering
    \includegraphics[width=0.9\linewidth]{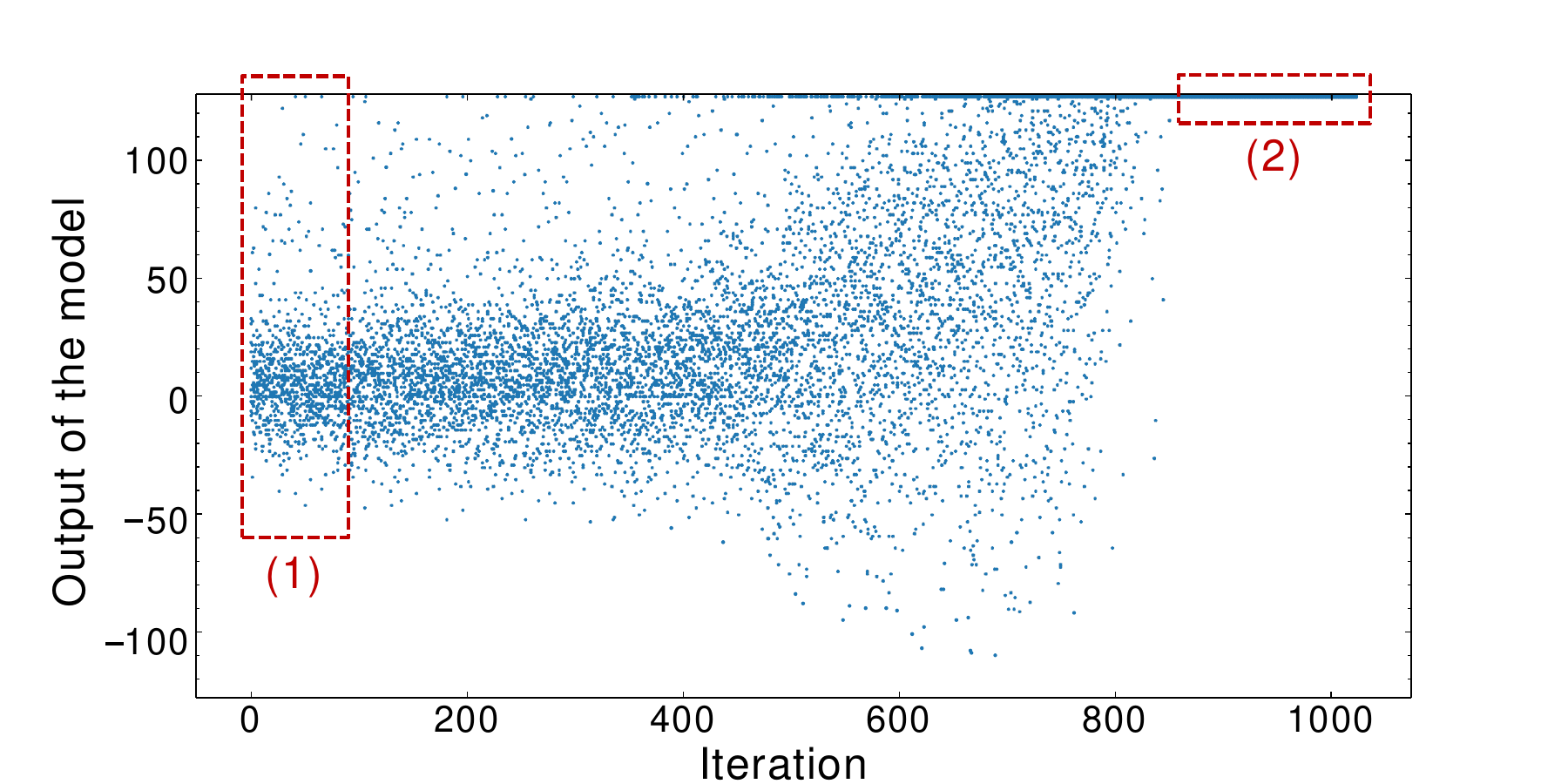}
    \caption{Transition of the output values of the model during the epoch when the accuracy drop happens in the existing integer-only training algorithm (NITI) with static scale factors. While the number of overflowed values ($\ge 127$) is almost zero at first (1), the output expands in the middle, and all values overflow by the end (2).}
    \label{fig/activation}
    \reducespace
\end{figure}

Therefore, this study focuses on static-scale integer-only training, which fixes all scale factors during on-device training and inference. Although some prior studies have performed neural network \textit{inference} with static-scale quantization\cite{staticscale1}\cite{hawqv3}, to the best of our knowledge, no studies have attempted static-scale integer-only neural network \textit{training}. Indeed, we empirically found that existing integer-only training methods with dynamic scaling are ineffective when replaced with static scaling. When a neural network was trained using the NITI algorithm with static scale factors, the accuracy suddenly dropped from 79\% to 11\% in the middle of the training. Figure \ref{fig/activation} shows the elements of the model output tensor for each input during the epoch when the accuracy drop occurred, demonstrating the explosion of the number of overflows in the output. We believe that the training collapsed because the model outputs became inaccurate due to overflow, resulting in inaccurate feedback to the model and leading to model weight updates in the wrong direction.

\reducespacebeforesection
\section{PRIOT: Pruning-Based Integer-Only Transfer Learning}\label{sec/priot}
\subsection{Design of PRIOT}\label{sec/priot-main}
To address the challenges in integer-only training, we propose PRIOT, a \underline{PR}uning-based \underline{I}nteger-\underline{O}nly \underline{T}ransfer learning. 
PRIOT freezes the pre-trained weights and optimizes the network by pruning the edges, training a pruning pattern to maximize accuracy. In PRIOT, the quantization scales do not change significantly because the weights are not updated and are only partially pruned. This stability prevents sudden drops in accuracy caused by inappropriate weight updates. Consequently, PRIOT avoids the training collapse that occurs during existing integer-only training with static scale factors.

For the pruning pattern training in PRIOT, we adopt the edge-popup algorithm originally proposed by Ramanujan et al. (2020) \cite{cvpr20_strong_lth}. This algorithm introduces a new parameter named \textit{score} assigned to each edge of the model, which is trained by backpropagation. Edges with low scores are pruned before inference. The forward pass is represented as follows:
\begin{align}
    \hat{W} &= W \odot \textit{mask}_p (S) \\
    y &= \hat{W}x
\end{align}
where $x$, $y$, $W$, and $S$ denote the input, output, weights, and scores of the layer, respectively, and $\odot$ represents element-wise multiplication. The $(i,j)$-th element of $\textit{mask}_p (S)$ is 1 when $(i, j)$-th element of $S$ is among the top $100-p\%$ elements of $S$ and otherwise 0. We skip the $\textit{mask}_p$ operation in the backward pass as it is not differentiable. This enables calculating the approximated score gradients $\delta S$ as follows:
\begin{align}
    \delta x &= \hat{W}^\top \delta y \label{eq/priot-bw}\\
    \delta S &= W \odot ((\delta y) x^\top)
\end{align}
where $\delta$ denotes the gradient of each tensor. 

Unlike the original edge-popup algorithm that uses randomly initialized weights, we fix weights to pre-trained values as our focus is on transfer learning. This maximizes the use of pre-trained information, achieving higher accuracy in shorter training time than training from scratch.

We also introduced two modifications to the edge-popup algorithm for lightweight computation. \underline{First}, we replace $\hat{W}$ in Equation \eqref{eq/priot-bw} with $W$ to reduce the computational overhead of masking $W$ in the backward pass, which we empirically verified to have little effect on the accuracy. \underline{Second}, whereas the original edge-popup algorithm fixes the pruning rate and selects the pruned edges by ranking the scores, our method introduces a hyper-parameter of the fixed score threshold and prunes the edges regardless of the pruning rate. This approach aims to avoid the computational cost of ranking all scores. 

\added{In this study, the scores for each layer are initialized with a normal distribution $\mathcal{N}(0, 32)$. Nevertheless, we have empirically found that the impact of the initialization method on accuracy is minimal, as the score updates in each training step are sufficiently large.}

\vspace{-1mm}
\subsection{PRIOT-S: Memory-Efficient Variant of PRIOT}\label{sec/priots}
Despite overcoming the challenge of static-scale integer-only training, PRIOT requires an increased memory footprint compared to ordinary training because scores must be stored in memory in addition to the original weights. 

To mitigate the increased memory footprint in PRIOT, we propose a memory-efficient variant named PRIOT-S, where S stands for sparsity. In PRIOT-S, a subset of edges is selected in advance, and scores are assigned only to these edges. Hence, less memory space is additionally required than the original PRIOT. Similar to PRIOT, edges with scores below the threshold are pruned before inference; thus, edges without scores are never pruned. When $M$ is a Boolean matrix representing the existence of scores, the forward pass of a single layer is expressed as follows: 
\begin{align}
    \hat{W} &= W \odot \textit{mask} (S, M) \\
    y &= \hat{W}x
\end{align}
where the $(i,j)$-th element of $\textit{mask} (S, M)$ is 1 if and only if $(i, j)$-th element of $M$ is 1 and $(i, j)$-th element of $S$ is not smaller than the threshold.

\deleted{In this study, we opt to randomly select scored edges, as we have not observed any improvement in accuracy when choosing edges based on specific factors, including weight values. We term the ratio of the unscored edges as the \textit{pruning rate} $p$, which is consistent across all layers. The pruning rate should be determined considering the trade-off between accuracy and computational costs.}
\added{Edges to be scored can be selected either randomly or heuristically based on metrics such as weights. While the latter may achieve higher accuracy, it comes with a trade-off of increased computational costs for initialization.}

\reducespacebeforesection
\section{Evaluation}\label{sec/eval}
\subsection{Setup}\label{sec/setup}
For the evaluation, we employ the Raspberry Pi Pico as the target device, which is a small microcontroller without an FPU. We target image classification tasks and design a tiny CNN model with two convolutional layers and two fully connected layers. The model is tailored to fit within the 264KB SRAM of the Raspberry Pi Pico. In addition to PRIOT and PRIOT-S, we implement NITI, an existing integer-only training algorithm, with static scale factors as a baseline; we call this \textit{static-scale NITI} hereafter. The quantization scheme in PRIOT and PRIOT-S is consistent with static-scale NITI. 

The model is first trained with a pre-training dataset on the host computer in an ordinary training manner using floating-point arithmetic. The pre-trained parameters of the model are then quantized and exported as global variables in the C++ implementation. The fixed scale factors are also calculated in this phase; we run quantized forward and backward passes with calibration data from the pre-training dataset, record the scale factor of each layer, and set each scale factor to the most frequent value. The model implementation is then compiled with the pre-trained parameters and static scale factors, and training is performed on the Raspberry Pi Pico with the target dataset. The batch size during training is set to 1.

We use the rotated MNIST dataset to evaluate the accuracy, as it is a popular benchmark for evaluating transfer learning \cite{rotatedmnist1}\cite{rotatedmnist2} and is feasible with tiny models like the one used in this study. Pre-training is conducted using the original MNIST dataset, where the model achieved 98.24\% top-1 test accuracy. Thereafter, on-device transfer learning is carried out using a subset of the MNIST dataset rotated by specific angles. We evaluate the accuracy with two rotation angles: 30{\textdegree} and 45{\textdegree}. Both on-device training and testing datasets consist of 1024 images each. Training is conducted for 30 epochs, and we evaluate the top-1 test accuracy using the model that achieved the highest top-1 training accuracy.
\added{Along with the evaluation on the Raspberry Pi Pico, we evaluated the accuracy of the rotated CIFAR-10 dataset using the VGG11 model to assess its generalizability to more complex tasks and larger models.}

\deleted{We evaluate PRIOT-S using two pruning rates: $90\%$ and $80\%$.}
\added{We evaluate four variants of PRIOT-S: two pruning rates of $90\%$ and $80\%$, and two methods for selecting scored edges, namely random selection and selecting edges with the largest absolute weights.}
The score threshold is consistent across all layers, set to $-64$ for PRIOT and $0$ for PRIOT-S.

\subsection{Results and Discussion}\label{sec/results}
Table \ref{table/acc} lists the best top-1 test accuracies during training using each method. In addition to static-scale integer-only training methods targeted in this study, we evaluate the original NITI with dynamic scale factors for reference. We conducted each experiment 10 times and calculated the mean and standard deviation, except for NITI and static-scale NITI, which offer no random factors in the experimental setup.

\begin{figure}
    \centering
    \includegraphics[width=0.7\linewidth]{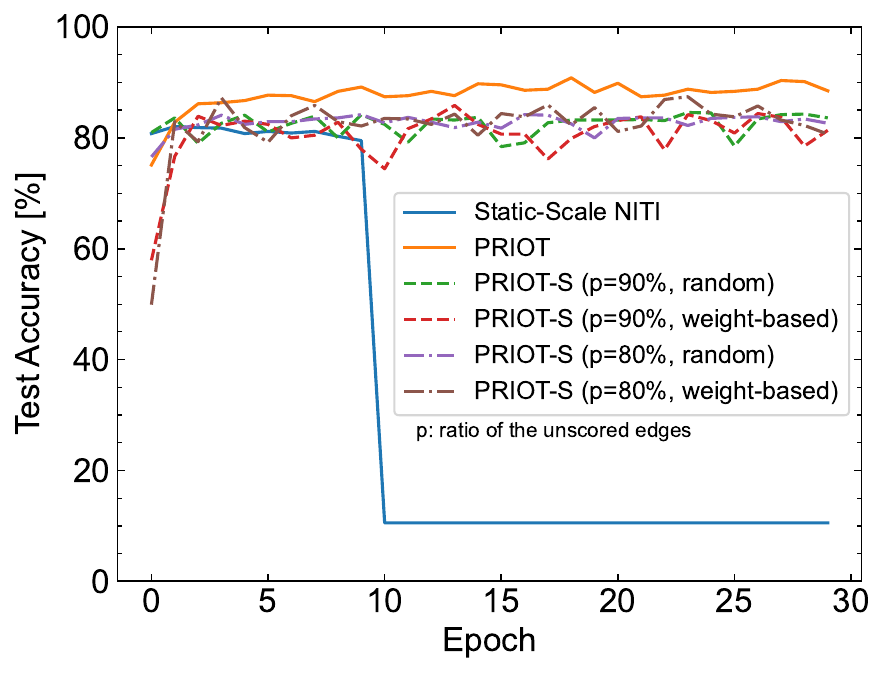}
    \caption{Accuracy history of each method with the rotated MNIST dataset with 30{\textdegree} rotation. While the accuracy of static-scale NITI drops in the middle, the accuracies of PRIOT and PRIOT-S continue to improve until the end.}
    \label{fig/priot-history}
    \reducespace
\end{figure}

While static-scale NITI struggles to achieve high accuracy, PRIOT demonstrates significant improvements, achieving an 8.08 percentage points (p.p.) accuracy improvement with a rotation angle of 30{\textdegree} and 33.75 p.p. with 45{\textdegree} rotations \added{in the rotated MNIST}, which are close to the reference dynamic-scale integer-only training.
\added{While achieving lower accuracy improvement than the rotated MNIST, PRIOT is also successful in training with the rotated CIFAR-10, showing its generalizability to more complex tasks.}
\deleted{For PRIOT-S, although the accuracy is lower than PRIOT, it still outperforms static-scale NITI by 3.01 p.p. and 14.69 p.p. with each dataset.}
\added{Although PRIOT-S achieves lower accuracy than PRIOT, it still outperforms static-scale NITI in most cases and remains effective in training, particularly when the scored edges are selected based on weights.}
The history of accuracies for each method is visualized in Figure \ref{fig/priot-history}, illustrating that the accuracy improvement achieved by PRIOT is attributed to the prevention of training collapse that occurs in static-scale NITI. 
\added{A further analysis on the distribution of scores and the pruned edges at every epoch shows that around 10\% of edges are pruned by the end in each layer. Although score variance grows over time, only a few edges fluctuate between pruned and unpruned, showing the stability of the training process.}

\begin{table}
    \centering
    \caption{Best top-1 accuracy during training with each method.}
    \begin{tabularx}{\hsize}{YY|YYY}
        \hline
        \multicolumn{2}{c|}{\added{Dataset}}& \multicolumn{2}{c}{MNIST} & \added{CIFAR-10} \\ 
        \multicolumn{2}{c|}{\added{Rotation Angles}}&  30{\textdegree} & 45{\textdegree} & \added{30{\textdegree}}\\ \hline\hline
        \multicolumn{2}{c|}{Before Transfer Learning} & 80.76 & 52.25 & \added{35.06} \\ 
        \multicolumn{2}{c|}{Dynamic-Scale NITI} & 90.43 & 90.72 &\added{38.57} \\ \hline
        \multicolumn{2}{c|}{Static-Scale NITI} & 80.86 & 51.95 &\added{35.06} \\ 
        \multicolumn{2}{c|}{\multirow{2}{*}{PRIOT}} & 88.94 $(\pm 1.02)$ & 85.70 $(\pm 1.03)$ &\added{55.16 $(\pm 1.05)$} \\ 
        \multirow{4}{*}{\parbox[t]{2cm}{PRIOT-S\\($p=90\%$)}} & \multirow{2}{*}{\added{random}} & \deleted{82.10 $(\pm 1.31)$} \added{80.35 $(\pm 2.86)$} & \deleted{58.85 $(\pm 2.54)$} \added{62.26 $(\pm 2.35)$} &\added{46.38 $(\pm 4.94)$} \\ 
        & \multirow{2}{*}{\added{weight-based}} & \added{80.05 $(\pm 3.53)$} & \added{75.76 $(\pm 3.34)$} &\added{10.74 $(\pm 0.00)$} \\ 
        \multirow{4}{*}{\parbox[t]{2cm}{PRIOT-S\\($p=80\%$)}} & \multirow{2}{*}{\added{random}} & \deleted{83.87 $(\pm 1.17)$} \added{82.81 $(\pm 1.85)$} & \deleted{66.64 $(\pm 1.54)$} \added{69.80 $(\pm 2.03)$} &\added{46.10 $(\pm 3.23)$} \\ 
        & \multirow{2}{*}{\added{weight-based}} & \added{83.12 $(\pm 1.67)$} & \added{82.05 $(\pm 1.72)$} &\added{10.74 $(\pm 0.00)$} \\ \hline
    \end{tabularx}
    \begin{tablenotes}[flushleft]
        \footnotesize
        \item While static-scale NITI, the existing method, shows almost no accuracy improvements from the pre-training, both PRIOT and PRIOT-S achieve accuracy improvements. In particular, PRIOT's accuracy improvement is significant and close to the reference dynamic-scale integer-only training. $p$ represents the ratio of the unscored edges in PRIOT-S. \added{The two rows of PRIOT-S correspond to the two methods of selecting scored edges.}
    \end{tablenotes}
    \label{table/acc}
    \reducespace
\end{table}

Table \ref{table/time} lists the training time and memory footprint during training for each method on the Raspberry Pi Pico. The training time refers to the time to run forward and backward passes for a single input image and is measured 100 times for each condition. For PRIOT, the training time increases by 4.13\% compared to static-scale NITI due to the on-the-fly pruning mask generation in the forward pass and increased computation in the score updates. In contrast, PRIOT-S achieves a decrease in computation time of 12.79\% compared to static-scale NITI. This reduction is attributed to the small number of parameter gradients to be calculated in PRIOT-S.

\begin{table}
    \centering
    \caption{Training time for a single input image and estimated memory footprint with each method on the Raspberry Pi Pico.}
    \begin{tabularx}{\hsize}{c|YY}
        \hline
          & \multicolumn{1}{c}{\multirow{2}{*}{Training Time [ms]}} & Estimated Memory Footprint [B] \\ \hline\hline
        Static-Scale NITI & 62.02 $(\pm 0.06)$ & 80,136 \\ 
        PRIOT & 64.58 $(\pm 0.08)$ & 138,044 \\ 
        PRIOT-S ($p=90\%$) & 52.77 $(\pm 0.05)$ & 97,672 \\ 
        PRIOT-S ($p=80\%$) & 54.09 $(\pm 0.09)$ & 102,880 \\ \hline
    \end{tabularx}
    \begin{tablenotes}[flushleft]
        \footnotesize
        \item While the computational costs of the PRIOT increase from the existing training algorithm (i.e., static-scale NITI), PRIOT-S requires much less computational costs than PRIOT, and its training time is even shorter than static-scale NITI. $p$ represents the ratio of the unscored edges in PRIOT-S.
    \end{tablenotes}
    \label{table/time}
    \reducespace
\end{table}

Regarding the memory footprint, we sum the sizes of the tensors stored during training, including activations, gradients, weights, and scores. Compared to static-scale NITI, PRIOT increases the memory footprint by 72.26\%, whereas PRIOT-S reduces it to 28.38\%. This result suggests that PRIOT-S effectively reduces the memory footprint when a certain level of accuracy loss is acceptable.
\added{Although PRIOT requires higher computational costs than PRIOT-S, it still significantly reduces the computational costs compared to dynamic-scale NITI and floating-point training, both of which cannot be executed on the Raspberry Pi Pico due to SRAM limitations.}

\reducespacebeforesection
\section{Conclusion}\label{sec/conclusion}
In this study, we introduced PRIOT, a pruning-based integer-only training method that enables effective training with static scaling factors. Additionally, we proposed a memory-saving variant, PRIOT-S. We implemented these algorithms on the Raspberry Pi Pico and evaluated their accuracy and computational costs. Our results show that PRIOT effectively prevents training collapse, which is observed in existing methods with static scale factors, and significantly improves accuracy. While PRIOT-S offers a smaller accuracy improvement than PRIOT, it substantially reduces computational costs, indicating its effectiveness in hardware-limited situations. While evaluated in limited situations in this study, we expect that our proposals will also be effective in other tasks and models, and valuable for a broad range of applications requiring efficient integer-only training on resource-constrained devices.
\bibliographystyle{IEEEtran}
\bibliography{main}

\end{document}